\documentclass{article}
\usepackage{graphicx} 
\usepackage[margin=1in]{geometry}

\usepackage{booktabs}
\usepackage{natbib} 
\usepackage{xcolor}
\usepackage{ulem}
\usepackage{hyperref}
\usepackage{amsmath}
\usepackage[utf8]{inputenc} 
\usepackage{textcomp} 
\usepackage{float}
\usepackage[affil-it]{authblk}
\usepackage[most]{tcolorbox}   
\usepackage{subcaption}        
\usepackage{caption}
\usepackage{tcolorbox}

\usepackage{graphicx}          
\tcbset{sharp corners=all, colback=white, colframe=black!15, boxrule=0.6pt}

\author[1,2]{Yazdan Babazadeh Maghsoodlo\thanks{Corresponding author: \texttt{ybabazad@uwaterloo.ca}}}

\author[2]{Madhur Anand}
\author[1]{Chris T. Bauch}
\affil[1]{Department of Applied Mathematics, University of Waterloo, Waterloo, Ontario, Canada,}
\affil[2]{School of Environmental Sciences, University of Guelph, Guelph, Ontario, Canada}

\title{Echoes Before Collapse: Deep Learning Detection of Flickering in Complex Systems}

\date{August 2025}

\begin{document}

\maketitle

\section{Abstract}


Deep learning offers powerful tools for anticipating tipping points in complex systems, yet its potential for detecting flickering (noise-driven switching between coexisting stable states) remains unexplored. Flickering is a hallmark of reduced resilience in climate systems, ecosystems, financial markets, and other systems.  It can precede critical regime shifts that are highly impactful but difficult to predict. Here we show that convolutional–long short-term memory (CNN–LSTM) models, trained on synthetic time series generated from simple polynomial functions with additive noise, can accurately identify flickering patterns. Despite being trained on simplified dynamics, our models generalize to diverse stochastic systems and reliably detect flickering in empirical datasets, including dormouse body temperature records and palaeoclimate proxies from the African Humid Period. These findings demonstrate that deep learning can extract early warning signals from noisy, nonlinear time series, providing a flexible framework for identifying instability across a wide range of dynamical systems.

\section{Introduction}

Sudden and irreversible shifts, tipping points \cite{russill2009tipping,evangelou2024tipping,he2024effect,luo1997bifurcation}, can arise in systems as varied as climate \cite{bury2019charting,lenton2000land}, ecosystems \cite{farahbakhsh2022modelling,jentsch2021fire,o2020tipping}, financial markets \cite{borah2025stability, kozlowska2016dynamic}, and public health \cite{WANG20161,kwuimy2020nonlinear,babazadeh2025big}. Anticipating such regime shifts is critical for mitigating their potentially severe consequences. Early warning signals (EWS) have emerged as an important tool for detecting the loss of resilience that precedes these critical transitions \cite{tredicce2004critical,bury2020detecting,dakos2012methods,dakos2013flickering,dylewsky2024early,dakos2023tipping,dakos2012robustness,marsden2012hopf,boettiger2012early,boettiger2012quantifying,lade2012early,drake2010early,kuehn2015early,guttal2008changing}.

One hallmark of reduced resilience is flickering: intermittent, noise-driven switching between coexisting stable states \cite{dakos2013flickering,kuehn2011mathematical,ashwin2012tipping,horsthemke1985noise}. Flickering has been documented in diverse systems, including palaeoclimate records \cite{trauth2024early,taylor1993flickering,ditlevsen1999observation,ditlevsen2010tipping}, neuroscience \cite{jercog2017up,holcman2006emergence}, hibernation physiology \cite{oro2021flickering}, and lake ecosystem dynamics \cite{wang2012flickering}. It often emerges before a bifurcation, when competing attractors become less stable and noise can drive rapid transitions between them \cite{kuehn2011mathematical,ditlevsen2010tipping}. This makes flickering a valuable EWS that can complement indicators based on critical slowing down. However, existing approaches to detecting flickering rely mainly on statistical metrics such as variance, autocorrelation, and skewness \cite{dakos2013flickering}. These methods can be sensitive to noise characteristics, data resolution, and confounding processes, limiting their robustness in practice \cite{dakos2013flickering,bury2021deep}.

In parallel, advances in artificial intelligence (AI) and deep learning (DL) have demonstrated considerable potential for identifying EWS of tipping points, particularly for detecting bifurcations in noisy, nonlinear systems \cite{bury2021deep,bury2023predicting,panahi2024machine,huang2024deep,deb2022machine,dylewsky2024early,dylewsky2023universal,math13172782}. Yet, despite its importance, the problem of detecting flickering using DL has not been addressed. This gap is significant: if DL models could reliably recognise flickering, they could expand our capacity to detect instability in a wider range of systems, especially when traditional statistical approaches fail.

Here, we develop a convolutional–long short-term memory (CNN–LSTM) framework to detect flickering patterns in noisy time series. We train the model on synthetic data generated from simple polynomial functions with additive noise, and test its generalization to a suite of more complex stochastic dynamical systems. We further show that the deep learning model consistently outperforms variance-based generic early warning indicators, achieving higher discrimination between true flickering and null cases. Finally, we apply the trained model to empirical datasets, including dormouse body temperature measurements \cite{oro2021flickering} and palaeoclimate records from the African Humid Period \cite{trauth2024early}. Our results show that deep learning can robustly identify flickering across both simulated and real-world systems, offering a new and broadly applicable tool for anticipating critical transitions.

\begin{tcolorbox}[colback=gray!3, title=Box 1. What is flickering?]
{\small
Flickering refers to noise-driven, intermittent transitions between coexisting stable states in a dynamical system. As a tipping point approaches, these switches become more frequent, indicating reduced resilience. (A–C) Potential landscapes V(x) as the control parameter p changes, altering the number and stability of equilibria. In (B), two nearby equilibria coexist, and noise drives rapid switching between them. (D) Corresponding time series x(t) with dashed lines marking the equilibrium positions from (A–C). The frequent switching in (B) is characteristic of flickering as an early warning signal.}\par\vspace{6pt}\hrule\vspace{8pt}

\begin{figure}[H] 
\captionsetup[subfigure]{labelformat=simple, labelsep=period, font=small}
\centering

\begin{subfigure}[t]{0.32\linewidth}
  \centering\includegraphics[width=\linewidth]{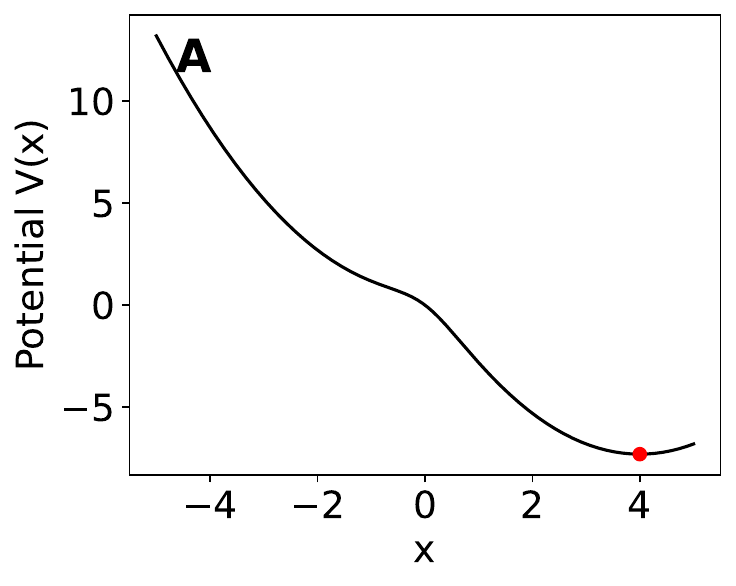}
\end{subfigure}\hfill
\begin{subfigure}[t]{0.32\linewidth}
  \centering\includegraphics[width=\linewidth]{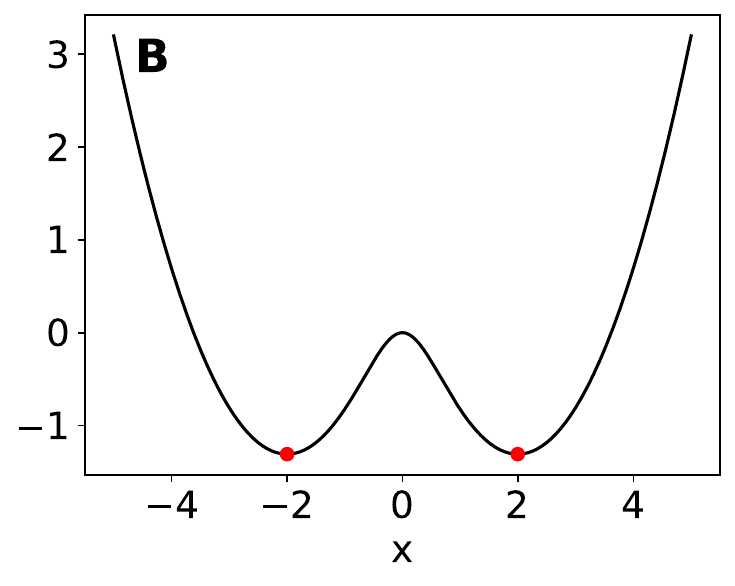}
\end{subfigure}\hfill
\begin{subfigure}[t]{0.32\linewidth}
  \centering\includegraphics[width=\linewidth]{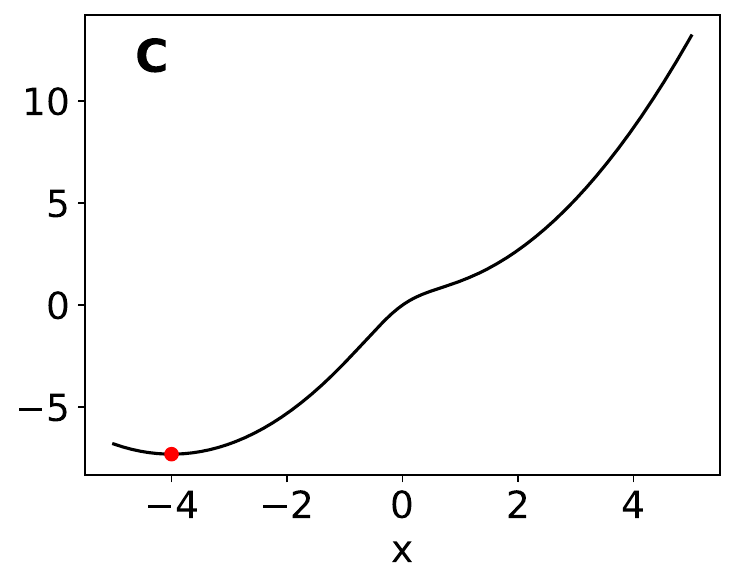}
\end{subfigure}

\vspace{8pt}

\begin{subfigure}[t]{0.80\linewidth}
  \centering\includegraphics[width=\linewidth]{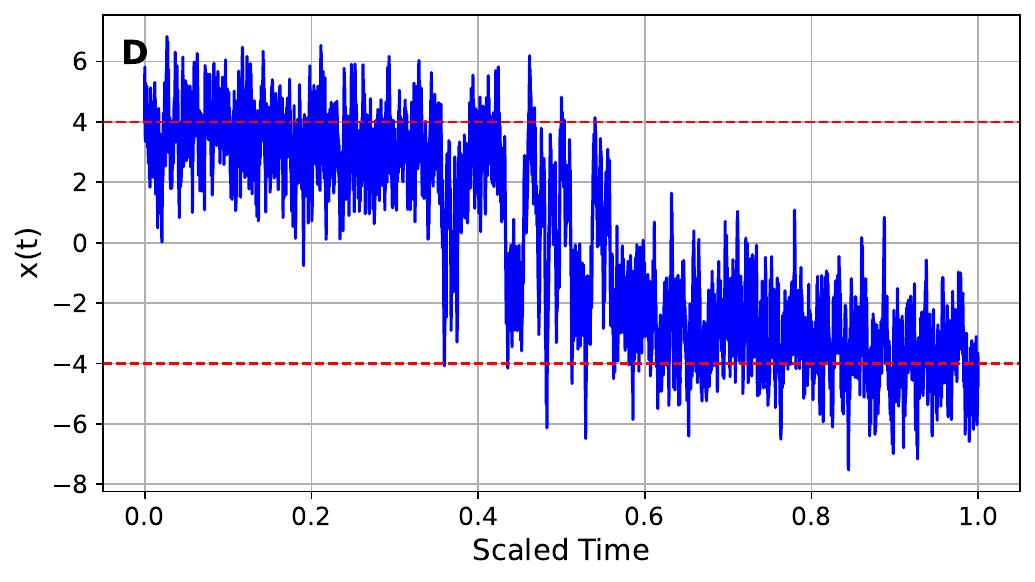}
\end{subfigure}

\end{figure}
\end{tcolorbox}

\section{Deep learning model and dataset construction}

\begin{figure}[H]
    \centering
    \includegraphics[width=\textwidth]{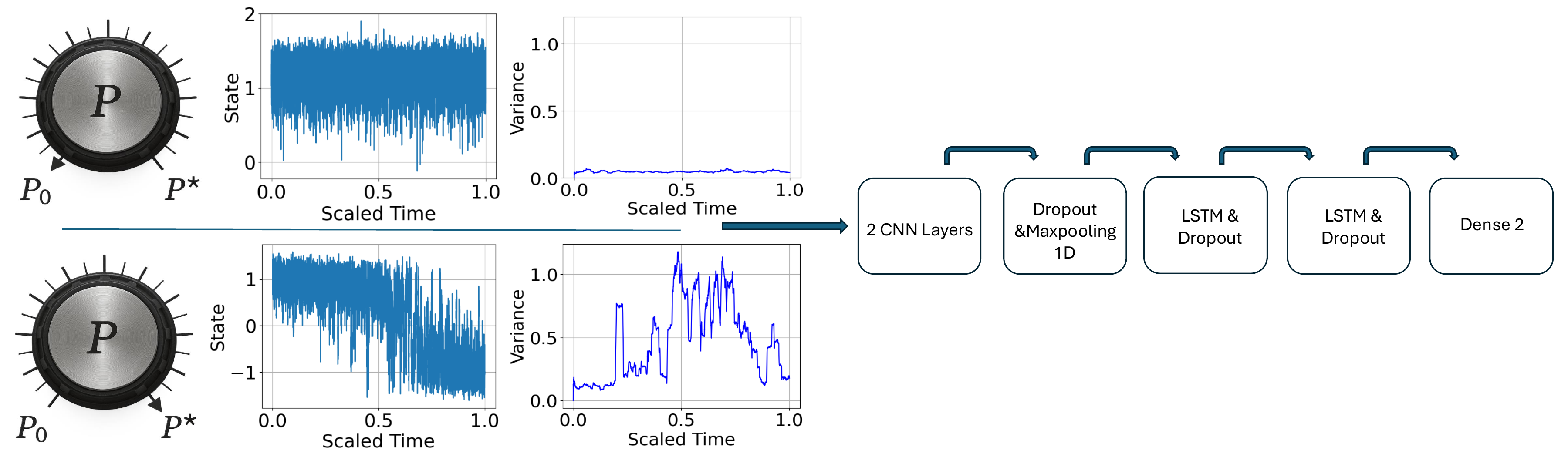}
    \caption{\textbf{Deep learning model architecture for flickering pattern detection} This figure shows the deep learning framework used to detect flickering patterns and provide early warnings of tipping points. Simulated time series with and without flickering ($p^*$, $p_0$) are pre-processed and used to train the model, which outputs a binary classification indicating the presence of flickering.}
    \label{fig1}
\end{figure}

We implemented a convolutional–long short-term memory (CNN–LSTM) network to detect flickering patterns in noisy time series \cite{sainath2015convolutional,alzubaidi2021review,hochreiter1997long}. The CNN layers extract localized features across long temporal windows, while the LSTM layers capture temporal dependencies and preserve relevant sequence information through recurrent connections. The final dense layer outputs a binary classification indicating the presence or absence of flickering. This architecture was chosen for its proven ability to combine spatial feature extraction with long-range temporal modeling in time series analysis \cite{dylewsky2025neural,dylewsky2024early, bury2023predicting, dylewsky2023universal, bury2021deep,math13172782}. A schematic of the architecture is shown in Fig.\ref{fig1}, and all layer specifications are provided in the Methods and Materials section.

Synthetic time series were generated from a one-dimensional stochastic differential equation consisting of a linear term, a seventh-degree polynomial, and additive Gaussian noise. The control parameter $p$ was varied to produce two categories:

\begin{itemize}
    \item \textbf{Flickering series}: $p$ was linearly varied from an initial value $p_0$ to the critical value $p^*$, producing intermittent switching between attractors near the transition.
    \item \textbf{Non-flickering series}: $p$ was fixed at $p_0$, keeping the system in a single attractor.
\end{itemize}

The noise amplitude $\sigma$ was tuned to balance state exploration and temporal structure preservation. Each time series was normalized using $z$-scores, and a rolling variance was computed in parallel to the raw signal. The final dataset comprised 2000 flickering and 2000 non-flickering series. Full equation parameters and coefficient ranges are detailed in the Methods and Materials.

Each sample was represented as a multivariate time series with two input channels (raw signal and rolling variance). We trained a series of CNN--LSTM models with identical architecture and hyperparameters but different sequence lengths $\{5000, 6000, 7000, 8000, 9000, 10000\}$. Each model was thus specialized to detect flickering patterns occurring over its respective temporal scale, with shorter windows capturing rapid local fluctuations and longer windows resolving extended pre-transition structures. Since genuine flickering requires a minimum temporal extent to manifest, overly short windows may miss the full pattern. All models achieved final training accuracies above $95\%$. Training across multiple scales and ensembling the outputs provides robust detection across the full range of temporal signatures. Full methodological details, including preprocessing, interpolation, and ensemble aggregation, are provided in the Methods and Materials.

\begin{figure}[H]
  \centering
  \includegraphics[width=\textwidth]{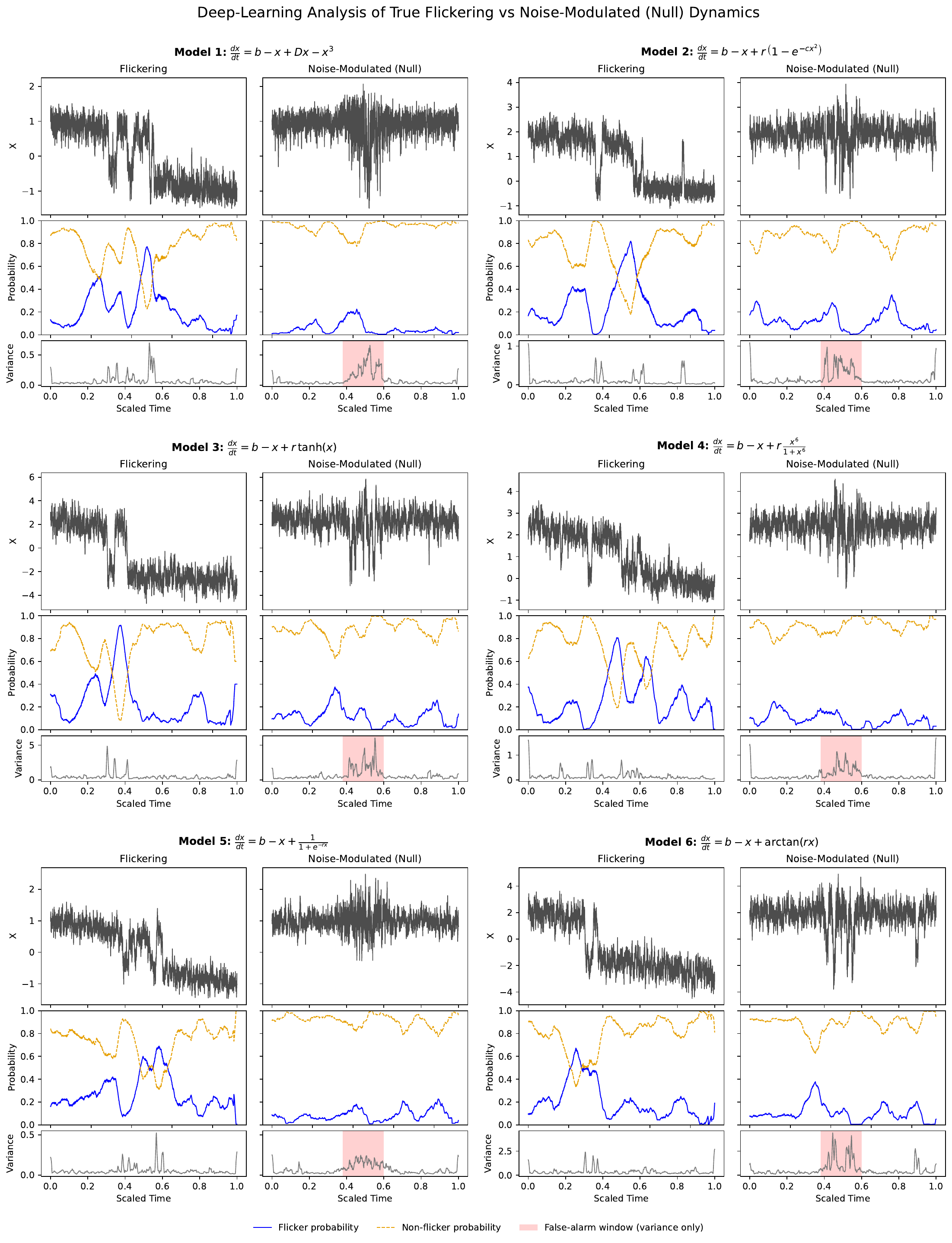}
  \caption{\textbf{Deep-learning discrimination of true flickering versus noise-modulated null dynamics across six models.} For each model, the left column depicts genuine flickering between coexisting equilibria, while the right column shows a null trajectory where the noise amplitude is modulated to mimic flicker. Within each panel, the top trace is the state time series, the middle trace shows the classifier’s flicker (blue) and non-flicker (yellow) probabilities, and the bottom trace shows the moving variance. The red band marks a noise-driven interval used to probe false-alarm behavior in the null case; axes are aligned to enable direct visual comparison.}
  \label{fig2}
\end{figure}

\section{Results}



We next tested whether the classifier detects flickering, and thereby offers a potential early-warning signal of an impending transition, across a suite of six nonlinear one-dimensional systems that were unseen during training and are structurally more complex than the synthetic time series used for model development (Fig. \ref{fig2}; model equations shown above each panel). For each system, we generated two classes of trajectories starting from a stable equilibrium under additive white noise. In the flickering condition (left column), we slowly advanced the control parameter toward the tipping point so that noise-induced switches between basins of attraction emerged prior to the transition. In the null condition (right column), the control parameter was held constant while the noise amplitude was exogenously increased and subsequently decreased over the middle third of the record to mimic a variance inflation that superficially resembles flickering but does not reflect proximity to a bifurcation. This null is intentionally challenging for generic indicators: the rolling variance (bottom row) rises in both conditions and would typically flag a false positive during the shaded ``false-alarm'' window. We passed each time series through an ensemble of deep learning detectors trained on different temporal contexts (window lengths equal to 8\%, 9.6\%, 11\%, 13\%, 14\%, and 16\% of the series), and averaged their posterior probabilities to obtain a single signal (middle row).

Across all six systems, the ensemble produced a sustained elevation of the ``flicker'' posterior that anticipates the critical transition in the flickering condition, while remaining quiescent in the null condition despite comparable variance inflation (Fig.~\ref{fig2}, middle vs bottom rows). In other words, when noise alone was modulated (null), the model correctly withheld an alarm even though classical variance-based EWS would have signalled during the shaded interval; by contrast, when genuine basin switching occurred (flickering), the model's probability trace rose clearly before the transition. These results demonstrate that the classifier generalises to novel, more complex dynamical systems not encountered during training, capturing dynamical features beyond low-order moment growth, and thereby discriminating true proximity to a tipping point from nuisance variance changes. Full simulation and training details, including noise schedules, parameter drifts, and ensemble aggregation, are provided in the methods and materials.

\begin{figure}[htbp]
    \centering
    \includegraphics[width=\textwidth]{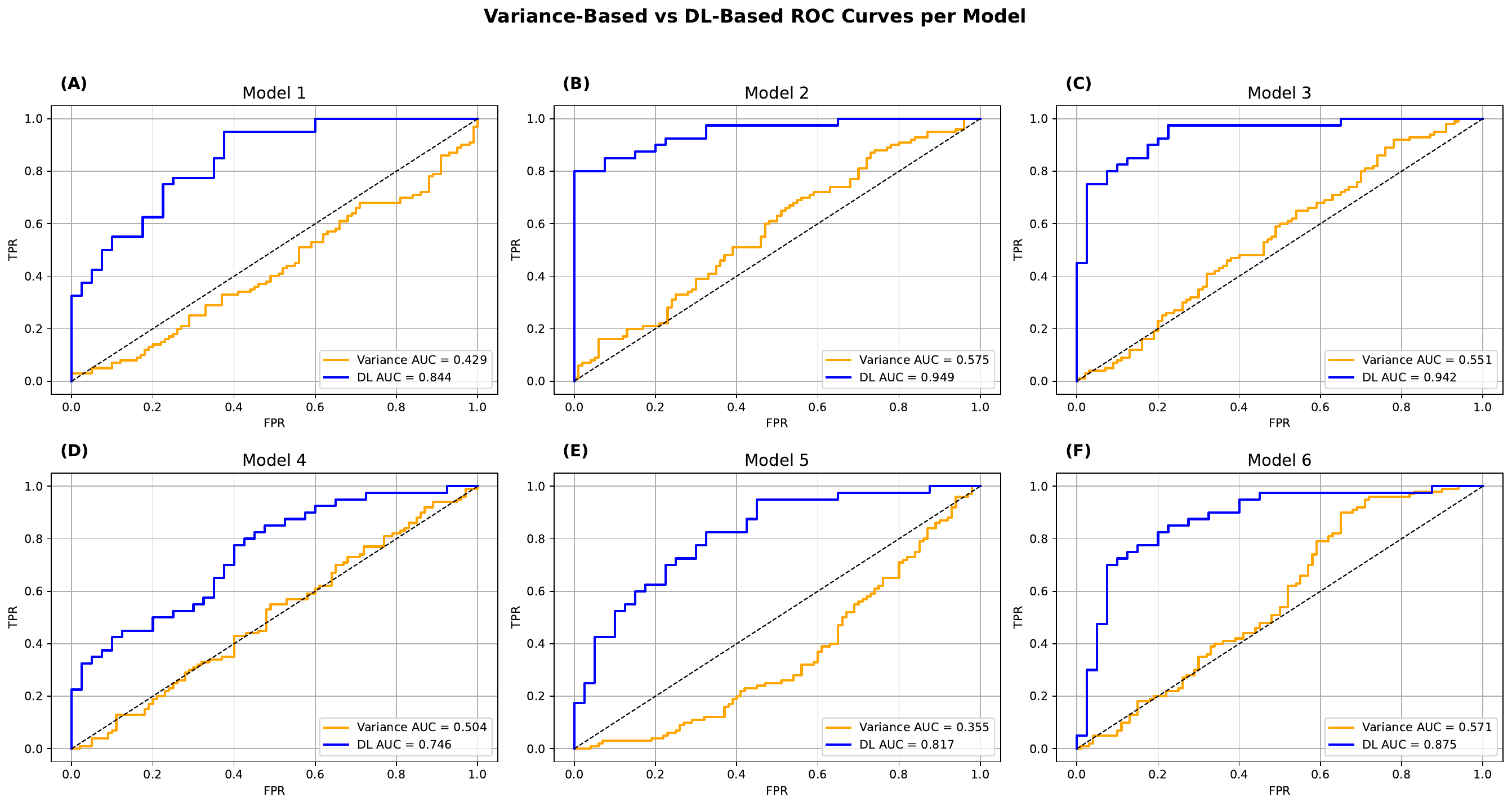}
\caption{\textbf{ROC performance on unseen dynamical systems.} 
Receiver operating characteristic (ROC) curves comparing a variance-based indicator (orange) and the deep learning (DL) classifier (blue) for six nonlinear systems (A–F). Curves are based on 300 simulations per system under flickering and null conditions. Variance-based AUCs hover near 0.5, indicating poor discrimination when variance inflates in both cases, while DL-based curves are consistently bowed toward the upper-left, yielding markedly higher AUCs and capturing flickering-specific dynamics beyond variance growth.}

    \label{fig3}
\end{figure}

To quantitatively assess how well each method distinguishes genuine flickering from the challenging null condition described above, we computed receiver operating characteristic (ROC) curves for both the variance-based indicator and the deep learning (DL) classifier on the same set of synthetic time series generated from the six unseen dynamical systems (Fig.~\ref{fig3}A--F). For each system, we conducted 300 independent stochastic simulations under both flickering and null conditions, following the same labelling protocol described in the preceding section. The variance-based indicator, which is historically used as a generic early warning signal, produced ROC curves that oscillate around the diagonal corresponding to a random classifier (AUC $\approx 0.5$) in most cases, confirming its inability to reliably distinguish between flickering and null scenarios when variance inflation occurs in both. In contrast, the DL classifier consistently produced ROC curves that bow sharply toward the upper-left corner, yielding substantially higher AUC values in every model tested. This consistent improvement demonstrates that the classifier can detect dynamical patterns characteristic of genuine flickering, beyond variance growth alone, and generalises effectively to dynamical systems that are both more complex and structurally distinct from those used during training.

\begin{figure}[htbp]
    \centering
    \includegraphics[width=\textwidth]{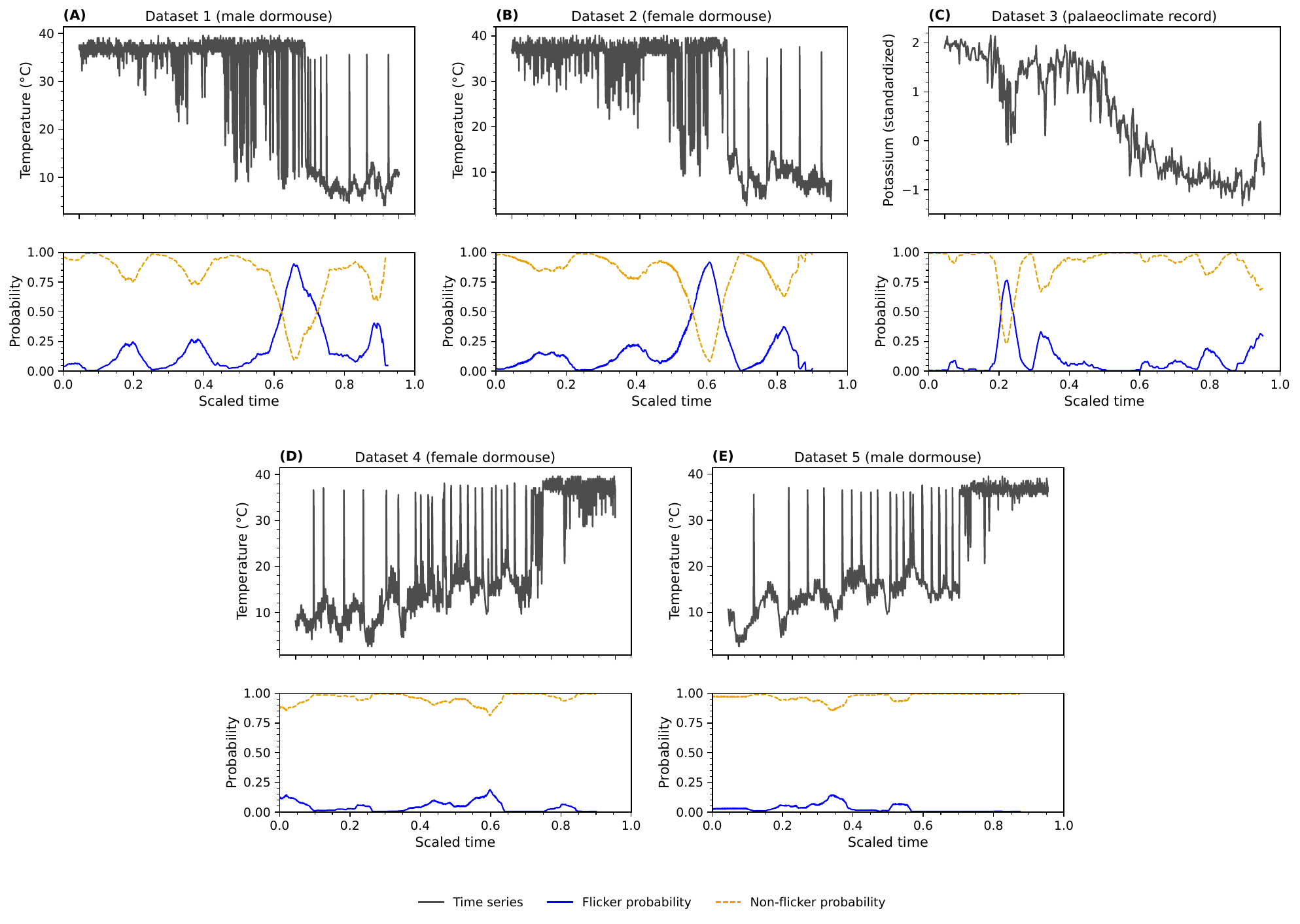}
\caption{\textbf{Application of the deep learning classifier to empirical datasets.} 
(A,B) Body temperature records from male and female edible dormice during the transition from active to hibernation states. (C) Standardised potassium concentrations from the Chew Bahir palaeoclimate record in southern Ethiopia, a proxy for aridity, spanning 9--3 kyr BP and capturing the termination of the African Humid Period. In each panel, the upper subfigure shows the original time series (Grey); the lower subfigure shows the deep learning ensemble probabilities for flickering (blue) and non-flickering (orange). Clear pre-transition flickering signals are detected in (A--C). (D,E) Two additional dormouse temperature records containing flickering that the classifier failed to detect, highlighting cases of missed detection. Probabilities are obtained by averaging outputs from models trained with sliding windows spanning 8\%, 9.6\%, 11\%, 13\%, 14\%, and 16\% of the total time series length.}

    \label{fig4}
\end{figure}

We further demonstrate that our method can be applied to empirical records, enabling early warning of real-world tipping points via detection of flickering patterns. We analysed two independent datasets. The first comprises high-resolution body temperature records from male and female edible dormice spanning several months as they enter and exit hibernation. In this dataset, clear flickering episodes occur shortly before the transition from active to hibernation states \cite{oro2021flickering}. The second dataset is a palaeoclimate record from the Chew Bahir basin in southern Ethiopia, covering 9--3 kyr BP, in which standardised potassium concentrations serve as a proxy for aridity. This record captures the abrupt termination of the African Humid Period, with a visible flickering phase preceding the shift to drier conditions \cite{trauth2024early}.  

Together, these datasets provide three empirical time series in which our deep learning ensemble reliably identifies flickering intervals and generates early warning signals prior to the observed transition points (Fig.~\ref{fig4}A--C). To harmonise temporal resolution, each series was upsampled to a standardised length via linear interpolation. Ensemble predictions were obtained by averaging the outputs of models trained with sliding windows spanning 8\%, 9.6\%, 11\%, 13\%, 14\%, and 16\% of the total series length.  In contrast, two additional time series from the same dormouse dataset, where pre-transition fluctuations were less pronounced, did not elicit an elevated flicker probability from the classifier (Fig.~\ref{fig4} D--E). We speculate that the algorithm training set would need to include features present in these time series (specifically, spacing between the temperature spikes) in order to be able to provide warning of a state change.  These results illustrate both the applicability and the specificity of the method: the model generalises to biological and palaeoclimate data while avoiding false positives in cases lacking clear dynamical signatures of flickering.

\section{Discussion}

We have demonstrated that deep learning models can provide reliable EWS of tipping points by detecting flickering patterns that often precede critical transitions. Our evaluation spanned three stages: (i) training on controlled synthetic datasets to capture essential flickering dynamics; (ii) testing on more complex stochastic models to assess generalisation; and (iii) applying the trained models to empirical datasets from biology and palaeoclimate. This pipeline demonstrates that deep learning can be trained on simplified representations yet retain predictive skill in diverse and realistic settings. Notably, the classifier maintained high performance even in challenging null scenarios where variance inflation alone could mislead classical EWS, highlighting its ability to extract dynamical features beyond low-order statistical moments.

The framework has several limitations. First, not all critical transitions are preceded by flickering, and the method will not issue warnings in such cases. Second, the training set relied on relatively simple functional forms, polynomials up to the seventh degree, limiting exposure to more intricate or high-dimensional behaviours. Third, the analysis focused on univariate time series, whereas many real-world systems are multivariate and feature interacting subsystems. Addressing these limitations will broaden the scope and applicability of the method.

Future work will expand the diversity of the training set to include a broader range of functional forms, noise structures, and dynamical regimes. Integrating the deep learning framework with complementary EWS approaches, such as indicators of critical slowing down, may improve detection in cases where flickering is absent or weak. Extending the approach to multivariate signals will enable application to complex systems where multiple interacting variables influence stability, such as coupled climate–biosphere systems or physiological networks.

Overall, our results establish a generalisable framework for machine-learning-based detection of flickering that moves beyond traditional variance-based EWS. By bridging simple synthetic training data with complex, unseen, and empirical systems, this approach offers a scalable and adaptable pathway for anticipating critical transitions in domains ranging from climate risk assessment and ecosystem management to biomedical monitoring and financial stability analysis.

\section{Methods and Materials}

\subsection{Details of Training Dataset}

To generate the training dataset, we simulated the following stochastic differential equation:

\begin{equation}
\frac{dx(t)}{dt} = p - x(t) + [ax(t) + bx^2(t) + cx^3(t) + dx^4(t) + ex^5(t) + fx^6(t) + gx^7(t)] + \sigma \xi(t),
\end{equation}

where \( \xi(t) \) denotes Gaussian white noise and \( \sigma \) is the noise intensity.

\paragraph{Random Coefficient Generation.} The coefficients of the polynomial term were randomly drawn from the following ranges:

\begin{itemize}
    \item \( g \in (0, -2) \)
    \item \( f \in (-g, +g) \)
    \item \( e \in (0, -2) \)
    \item \( d \in (-e, +e) \)
    \item \( c \in (0, -2) \)
    \item \( b = 0 \)
    \item \( a \in (1, 3) \)
\end{itemize}

\paragraph{Initial Conditions and Noise.} We set \( p = 5 \), referred to as \( p_0 \), to ensure a sufficient distance from the transition point. The positive root of the right-hand side of equation (4) was used as the initial condition \( x_0 \). The noise intensity was set as \( \sigma = 1.2 x_0 \), which ensures the emergence of flickering behaviour near the transition.

\paragraph{Determination of Transition Point.} We identified the value \( x^* \) where the slope of the polynomial term equals that of the linear term (i.e., slope = 1). The corresponding critical value \( p^* \) was then computed using:

\begin{equation}
y^* = x^* - p^*.
\end{equation}

\paragraph{Simulation Procedure.} Using the Euler–Maruyama method, we generated time series of lengths 5000, 6000, 7000, 8000, 9000, and 10,000 steps. For time series with flickering, the parameter \( p \) was linearly varied from \( p_0 \) to \( p^* \). For non-flickering series, \( p \) remained fixed at \( p_0 \).

\paragraph{Dataset Composition.} A total of 2000 time series were generated for each category (flickering and non-flickering). For each series, we computed its rolling variance using a window size of 1000, and both the raw time series and the rolling variance (each followed by z-score normalisation) were used as the two input channels to the deep learning model.

\subsection{Deep Learning Architecture}

The deep learning model used in this study comprises three main components: convolutional feature extraction, temporal sequence modelling, and classification. The model was implemented in Keras using the TensorFlow backend, and designed to capture both spatial patterns in the input time series and their temporal evolution.

\paragraph{Feature Extraction:}
To extract localized patterns across long temporal windows, the model starts with two 1D convolutional layers using large kernels. Specifically:
\begin{itemize}
    \item \textbf{Conv1D layer 1:} 50 filters, kernel size = 300, activation = ReLU, padding = 'same'
    \item \textbf{Conv1D layer 2:} 100 filters, kernel size = 300, activation = ReLU, padding = 'same'
    \item \textbf{Dropout:} dropout rate = 0.05
    \item \textbf{MaxPooling1D:} pool size = 2, stride = 2, padding = 'valid'
\end{itemize}

\paragraph{Temporal Modelling:}
The temporally encoded features are passed through two stacked LSTM layers:
\begin{itemize}
    \item \textbf{LSTM layer 1:} 50 memory cells, return\_sequences = True
    \item \textbf{Dropout:} dropout rate = 0.05
    \item \textbf{LSTM layer 2:} 10 memory cells
    \item \textbf{Dropout:} dropout rate = 0.05
\end{itemize}


\paragraph{Classification:}
The final hidden representation is passed to a fully connected output layer:
\begin{itemize}
    \item \textbf{Dense layer:} 2 units, activation = softmax
\end{itemize}
This outputs the class probabilities for binary classification.

\paragraph{Training Configuration:}
The model was compiled with the Adam optimizer with a learning rate of 0.01 and trained using the sparse categorical cross-entropy loss. A batch size of 32 was used. To prevent overfitting, early stopping with a patience of 2 epochs and model checkpointing based on validation accuracy were applied. The model was trained for up to 5 epochs, and the best-performing model (based on validation accuracy) was saved. The input shape for each sample was \texttt{(5000, 2)}, corresponding to 5000 time steps with 2 features per time step.

This configuration enabled the model to achieve 100\% validation accuracy within 5 epochs on the given training and validation data.

\subsection{Details of Simulated Models}

To evaluate the performance of our trained deep learning model, we tested it on six distinct nonlinear stochastic dynamical systems. For each model, the simulation was run for 62,500 time steps using the Euler–Maruyama method with a fixed time step $\Delta t = 0.01$. All models incorporate a stochastic forcing term $ \sigma \xi(t) $, where $ \xi(t) $ represents Gaussian white noise.  

Two simulation regimes were implemented for each model:  
(1) \textbf{Flickering case}: a control parameter ($b$) was varied linearly over time from an initial to a final value, reducing the stability of one attractor and inducing noise-driven switching between alternative states prior to the bifurcation point, while the noise amplitude $\sigma$ was held constant.  
(2) \textbf{Non-flickering case}: the control parameter $b$ was held constant, but the noise amplitude $\sigma$ was transiently increased in a triangular profile during the middle third of the simulation, producing variance inflation without state transitions.  

This design ensured that flickering and non-flickering time series exhibited similar variance dynamics while differing fundamentally in their underlying attractor-switching behaviour, making discrimination nontrivial.

The six models and their parameter settings were as follows:

\begin{enumerate}
    \item \textbf{Cubic Polynomial Model:}
    \begin{equation}
    \frac{dx}{dt} = b - x + Dx - x^3 + \sigma \xi(t)
    \end{equation}
    with $D = 1.5$, $\sigma = 0.4$, and $b$ starting from $0.5$ and ramping to $b_{\mathrm{final}}$ in the flickering case.

    \item \textbf{Exponential Potential Model:}
    \begin{equation}
    \frac{dx}{dt} = b - x + 2\left(1 - e^{-2x^2}\right) + \sigma \xi(t)
    \end{equation}
    with $b$ decreasing from $0$ to $-1$ in the flickering case and $\sigma = 0.45$.

    \item \textbf{Hyperbolic Tangent Model:}
    \begin{equation}
    \frac{dx}{dt} = b - x + 2 \tanh(x) + \sigma \xi(t)
    \end{equation}
    with $b$ decreasing from $0.5$ to $-1$ in the flickering case and $\sigma = 0.9$.

    \item \textbf{Sigmoidal Hill Function Model:}
    \begin{equation}
    \frac{dx}{dt} = b - x + 1.5 \frac{x^6}{1 + x^6} + \sigma \xi(t)
    \end{equation}
    with $b$ decreasing from $1$ to $-0.5$ in the flickering case and $\sigma = 0.5$.

    \item \textbf{Logistic Function Model:}
    \begin{equation}
    \frac{dx}{dt} = b - x + \frac{1}{1 + e^{-10x}} + \sigma \xi(t)
    \end{equation}
    with $b$ decreasing from $0$ to $-1$ in the flickering case and $\sigma = 0.3$.

    \item \textbf{Arctangent Potential Model:}
    \begin{equation}
    \frac{dx}{dt} = b - x + \arctan(10x) + \sigma \xi(t)
    \end{equation}
    with $b$ decreasing from $0.5$ to $-1$ in the flickering case and $\sigma = 0.8$.
\end{enumerate}

\subsection*{ROC curve construction}

We evaluated discrimination between flickering(positive) and null (negative) trajectories using receiver operating characteristic (ROC) analysis.  
For the DL detector, we generated 50 trajectories per dynamical model for each class.  
The DL detector consisted of an ensemble of CNN–LSTM networks trained on window lengths of 5{,}000, 6{,}000, 7{,}000, 8{,}000, 9{,}000, and 10{,}000 samples.  
For each trajectory, we computed the scalar score  
\[
s_{\mathrm{DL}} = \max_{w} p_{\mathrm{f}}(w) - \mathrm{mean}_{w}[p_{\mathrm{f}}(w)],
\]  
where $p_{\mathrm{f}}(w)$ is the per-window flicker probability output by the DL model.  
To obtain a conservative estimate, the weakest score among the six DL models (i.e., the minimum $s_{\mathrm{DL}}$ across lengths) was retained for both flickering and null cases.  
The subtraction of the mean accounts for the fact that $p_{\mathrm{f}}$ is already normalized to $[0,1]$.

For the variance-based baseline, we generated 100 trajectories per model for each class.  
We computed the rolling variance $V(t)$ with a window of 1{,}000 samples, and defined the scalar score as  
\[
s_{\mathrm{var}} = \frac{\max_{t} V(t)}{\mathrm{mean}_{t}[V(t)] + \mathrm{std}_{t}[V(t)]}.
\]  
The denominator normalizes the maximum variance surge by the baseline mean and variability, making it interpretable despite the unscaled nature of $V(t)$.

In both cases, ROC curves were obtained by sweeping a classification threshold $\tau$ over the sorted set of scores $\{s_i\}$ and computing true- and false-positive rates at each $\tau$.  
The area under the curve (AUC) was calculated using trapezoidal integration.  
All ROC analyses were performed separately for each dynamical model.

\subsection{Details of Empirical Data}

The first set of empirical data used in this analysis consists of body temperature measurements from male and female edible dormice, recorded under Mediterranean outdoor conditions. The animals were monitored continuously over the course of nearly one year, with temperature loggers implanted to measure hourly body temperature alongside ambient air temperature. The data exhibit two distinct regimes corresponding to active and hibernation phases, separated by abrupt transitions. During hibernation, both individuals displayed frequent and irregular alternations between torpid and euthermic states—indicative of flickering dynamics—which intensified as the transition to the active phase approached. In contrast, body temperature remained relatively stable during active periods. Notably, the female entered hibernation approximately three weeks earlier than the male, although both resumed activity on the same day, suggesting both individual variation and synchronised physiological cues. These observations provide empirical support for the interpretation of flickering as a potential early warning signal of critical physiological transitions.

The second dataset employed in this study is a high-resolution palaeoclimate record from the Chew Bahir basin in southern Ethiopia, spanning the termination of the African Humid Period (AHP) between 9 and 3 thousand years before present (kyr BP). This dataset is based on potassium concentrations in lake sediments, a well-established proxy for regional aridity. The record reveals a sequence of approximately fourteen drought events, each lasting 20–80 years and recurring at intervals of around 160 years. These droughts precede and accompany a ~1000-year transition from a stable wet to a stable dry climate state. The recurrence and clustering of short drought and wet events generate a distinct pattern of climate flickering—rapid alternations between opposing climate states—that becomes more pronounced as the system nears the critical transition. These findings offer compelling evidence that flickering can act as an early warning indicator of impending tipping points in large-scale climate systems (Trauth et al., 2024).

To ensure consistency in model input dimensions, all empirical time series were interpolated linearly to a uniform length of 100,000 time steps.

\section{Data Availability}

All the codes and empirical data used in this research can be found in the following repository:

\url{https://github.com/Yazdan-Babazadeh/Echoes-Before-Collapse}


\end{document}